# Syllabification by Phone Categorization

Jacob Krantz, Maxwell Dulin, Paul De Palma
Gonzaga University
Spokane, WA
jkrantz@zagmail.gonzaga.edu

Mark VanDam
Washington State University
Spokane, WA
mark.vandam@wsu.edu

## ABSTRACT
Syllables play an important role in speech synthesis, speech recognition, and spoken document retrieval. A novel, low cost, and language agnostic approach to dividing words into their corresponding syllables is presented. A hybrid genetic algorithm constructs a categorization of phones optimized for syllabification. This categorization is used on top of a hidden Markov model sequence classifier to find syllable boundaries. The technique shows promising preliminary results when trained and tested on English words.

## 1 INTRODUCTION

Syllables play an important role in speech synthesis, speech recognition, and spoken document retrieval. Speaking broadly, words are composed of syllables, which are composed of phones, where a phone is a unit of sound, like [t] in the English tip. The traditional approach to syllabification has been to characterize the syllable as an optional consonant (C) onset, followed by a vowel (V) nucleus, followed by an optional consonant coda, with sound rising from the onset to the nucleus and falling toward the coda. All languages appear to have at least {V, CV} in their syllable inventories. Many supplement these with codas, including {VC, CVC} syllable types.

Surveys of automatic syllabification research are provided in [3] and [4]. Approaches are either rule-based, which presume an underlying theory of syllabification, or data driven which infer new syllabifications from corpora of words assumed to be correctly syllabified. Data-driven syllabification appears to perform much better than rule-driven methods. The syllabifier developed at the National Institute of Standards and Technology (NIST), based on Daniel Kahn's 1979 MIT dissertation, is an example of a rule-based approach. It was widely used for at least a decade [1], [2]. Data driven approaches include look-up procedures using weighted finite state automata, and various machine-learning techniques (ML), with a back-propagation result dating from 1992. The ML techniques achieve impressive accuracy, though at the cost of great complexity. For example, one syllabifier uses a fifth order hidden Markov model, another uses a hybrid support vector machine in tandem with a hidden Markov model. Neither appears to be publicly available, either for research use or for replication [3], [4].

Most discussions of syllabification treat it as a high-level computational linguistics problem with no mention of underlying hardware or programming techniques. This is unfortunate, since inferences from very large data sets are compute-intensive. The work presented in this paper achieved a 33% decrease in turn-around time after using Amazon Web Services (AWS) hardware enhanced by parallelization. These have allowed training on a larger corpus and, because of reduced turn-around time, provided more opportunities to refine the software. This paper is a preliminary report on research whose goals are to build an efficient, accurate, relatively simple, and cross-linguistic syllabifier. The software itself, along with the data, will be made available to other investigators, both for replication and research.

## 2 METHOD

The syllabification of written words can be treated as a sequence classification problem. The presented approach combines classification by a first-order hidden Markov model (HMM) with optimization from the genetic algorithm. The input to the syllabification model is a word represented as a sequence of phones in DISC format, a variation of the International Phonetic Alphabet (IPA). Before classification, the input phones undergo a transformation where each phone in the sequence is replaced with the category it belongs to. A table is maintained by the system that contains a set of many-to-one mappings of phones to categories. For example, the following are phone-category mappings, where the phones on the left map to a category on the right:

$$b \rightarrow b \qquad \{, E \rightarrow a \qquad s, n, t \rightarrow c$$

Using the above mappings, the word absent [{bsEnt] is mapped to the sequence {a, b, c, a, c, c}, where each element of the sequence is an arbitrary name representing a distinct phone category. This category sequence, enumerated as bigrams, forms the observation sequence sent to the HMM. Given the observation sequence and a trained HMM, the Viterbi algorithm determines the most likely locations of syllable boundaries. Using these syllable boundary locations and the original input word, a final syllabification can be synthesized.

Perplexity is a measurement that quantifies the difficulty of predicting the next token of a sequence and is related to the entropy of a model's training data [5]. For an HMM, the hidden state space is representative of the overall perplexity. By using phone-category mappings, the system has reduced the hidden state space from 108 states down to 24, thus reducing the perplexity. A lower perplexity makes the HMM more tractable. Beyond tractability, mapping phones to categories increases the knowledge of the model by associating like phones that similarly affect syllabification.

For the model to be language agnostic, the training process is two-fold: 1) condition the HMM on syllabified examples and 2) generate the phone-category mappings. Supervised training of the HMM is accomplished by extracting words annotated with syllabic boundaries from the CELEX2 lexical dataset [6]. From each training word, both an observation sequence and a hidden state sequence are generated. Using the training example of the disyllabic word "ab-sent" [{b-sEnt], the observation sequence becomes $o = [ab, bc, ca, ac, cc]$ while the corresponding hidden state sequence becomes $s = [a0b, b1c, c0a, a0c, c0c]$. The observation sequence is the bigram enumeration of phone-category mappings. Each hidden state relates to observation. The hidden states thus denote whether or not a syllable boundary exists between the two categories of the bigram.

Generation of the phone-category mappings is the second training process. Conventional[1] phone categories with the HMM described above produces an 83.45% accuracy. With dozens of distinct phones, the determination of the space of candidate phone-category mappings is large. This problem naturally lends itself to optimization through the genetic algorithm. The algorithm is initially presented a set of randomly generated phone-category mappings whose fitness is evolutionarily judged by the accuracy of the mappings in producing syllabifications. The stochastic universal sampling method (SUS) is used to select a population for mating. SUS allows for a potentially diverse mating population while weighting most heavily those candidates with highest fitness values. Mating is achieved through scattered crossover. This ensures that child 1 will have approximately half of parent 0's and parent 1's genes, or phones-category mappings. Child 2 will then be the exact opposite of child 1, in order to preserve all of the phone-category mappings from the parents. The mutation rate is dynamically adaptive depending on the fitness of the population, higher if the values are close together, lower if the values are dispersed. The dynamic technique promotes more variation among chromosomes than a static mutation rate would. A final fine-grained bit of optimization is performed, periodically, by observing the gene, or phone, that is most frequently involved in mis-syllabifications in the most fit member of the population. The HMM is trained and tested with the phone permuted in every possible categorization to determine which categorization performs best for the given phone. The highest fitness replaces the most fit of the population.

---

[1] By "conventional", the authors intend the categories found in the International Phonetic Alphabet chart (http://www.internationalphoneticalphabet.org/ipa-sounds/ipa-chart-with-sounds/, retrieved 2/4/2018).

## 3 RESULTS AND DISCUSSION

Three major results have been yielded at different steps in the project. First, results were calculated for an isolated first order HMM at 73.20% accuracy. Phone categorizations were then added using conventional linguistic knowledge boosting accuracy to 83.45%. Finally, the phone categorizations were generalized and optimized with the genetic algorithm, resulting in a syllabification accuracy of 92.54%. The genetic algorithm increases accuracy sharply in early generations, with the top scheme surpassing itself 16 times in the first hundred generations. It beats both the no category scheme and the conventional category scheme within 40 generations. Accuracies were determined using 10-fold cross validation on a set of 60K examples.

## 4 FUTURE WORK

Currently, the system implements a first order HMM and is constrained to 12 different phonetic categories. To improve the model, the genetic algorithm will be run for all potential numbers of phonetic categories. Increasing the order of the HMM may also increase the accuracy threshold. Once accuracy has been maximized in English, language-independence will be tested against German, Dutch, and other languages. Improved hardware is necessary to facilitate training of the model, requiring a larger AWS instance in the future. An interesting result of the genetic algorithm is the generated phone-categories, which do not appear to pattern well with conventional natural classes of phones. A further investigation will analyze why certain phones are shown to act similarly in syllabification. Finally, since there is no agreed-upon theory of syllabification, future work will use other syllabified lexicons for training. The data, along with the system of syllabification, will be made publicly available.